\documentclass{article}

\usepackage[preprint]{neurips_2024}

\usepackage[utf8]{inputenc} % allow utf-8 input
\usepackage[T1]{fontenc}    % use 8-bit T1 fonts
\usepackage{hyperref}       % hyperlinks
\usepackage{url}            % simple URL typesetting
\usepackage{booktabs}       % professional-quality tables
\usepackage{amsfonts}       % blackboard math symbols
\usepackage{nicefrac}       % compact symbols for 1/2, etc.
\usepackage{microtype}      % microtypography
\usepackage{xcolor}         % colors
\usepackage{multirow}

\usepackage{paralist}
\usepackage{booktabs}
\usepackage{multicol}
\usepackage{multirow}
\usepackage{bm}
\usepackage{bbm}
\usepackage{diagbox}
\usepackage{makecell}
\usepackage{amssymb}
\usepackage{amsmath}
\usepackage{amsthm}
\usepackage{booktabs}
\usepackage{enumitem}
\usepackage{graphicx}
\usepackage{color}

\title{Training-Free Open-Ended Object Detection and Segmentation via Attention as Prompts}

\author{%
  Zhiwei Lin \quad 
  Yongtao Wang \thanks{Corresponding author} \quad 
  Zhi Tang \\
  Wangxuan Institute of Computer Technology, Peking University, China\\
  \texttt{\{zwlin, wyt, tangzhi\}@pku.edu.cn} \\
}

\begin{document}

\maketitle

\begin{abstract}
Existing perception models achieve great success by learning from large amounts of labeled data, but they still struggle with open-world scenarios.
%
% To alleviate this issue, open-set perception models are proposed. 
To alleviate this issue, researchers introduce open-set perception tasks to detect or segment unseen objects in the training set. 
%
% However, these models rely on object categories as inputs, which are not available in real-world scenes, or require foreground object proposals, which also depend on training data and face challenges when unseen objects are encountered. 
% However, these models rely on object categories as inputs, which are not available in real-world scenes.
However, these models require predefined object categories as inputs during inference, which are not available in real-world scenarios.
Recently, researchers pose a new and more practical problem, \textit{i.e.}, open-ended object detection, which discovers unseen objects without any object categories as inputs.
%
% In this paper, we present VL-SAM, a framework that combines the generalized object recognition model (\textit{i.e.,} Vision-Language Model) with the generalized object localization model (\textit{i.e.,} Segment-Anything Model), to address the object detection task for corner cases. 
In this paper, we present VL-SAM, a training-free framework that combines the generalized object recognition model (\textit{i.e.,} Vision-Language Model) with the generalized object localization model (\textit{i.e.,} Segment-Anything Model), to address the open-ended object detection and segmentation task. 
Without additional training, we connect these two generalized models with attention maps as the prompts.
%
% Specifically, we employ a regularized attention flow mechanism to propagate attention maps across all layers and heads in VLM, yielding high-quality attention maps. 
Specifically, we design an attention map generation module by employing head aggregation and a regularized attention flow to aggregate and propagate attention maps across all heads and layers in VLM, yielding high-quality attention maps. 
Then, we iteratively sample positive and negative points from the attention maps with a prompt generation module and send the sampled points to SAM to segment corresponding objects. 
%
% Experimental results on the corner case object detection dataset (CODA) and long-tail instance segmentation dataset (LVIS) show that our method achieves competitive performance compared to existing models, demonstrating the effectiveness of VL-SAM. 
Experimental results on the long-tail instance segmentation dataset (LVIS) show that our method surpasses the previous open-ended method on the object detection task and can provide additional instance segmentation masks.
% Meanwhile, achieves competitive performance compared to open-set models.
Besides, VL-SAM achieves favorable performance on the corner case object detection dataset (CODA), demonstrating the effectiveness of VL-SAM in real-world applications. 
Moreover, VL-SAM exhibits good model generalization that can incorporate various VLMs and SAMs.
% The source code will be made publicly available.

\end{abstract}

% \section{Submission of papers to NeurIPS 2024}
\section{Introduction}
% Deep learning has achieved remarkable success in autonomous driving perception tasks.
Deep learning has achieved remarkable success in perception tasks, with autonomous driving as a typical practical application.
Existing deep learning based perception models rely on extensive labeled training data to learn to recognize and locate objects. 
However, training data cannot cover all types of objects in real-world scenarios. 
%
% When faced with out-of-distribution objects, \textit{i.e.}, corner cases, existing perception models may fail to recognize and locate objects, leading to severe safety issues~\cite{li2022coda}.
% For instance, in autonomous driving scenarios, when faced with out-of-distribution objects, existing perception models may fail to recognize and locate objects, leading to severe safety issues~\cite{li2022coda}.
When faced with out-of-distribution objects, existing perception models may fail to recognize and locate objects, which can lead to severe safety issues~\cite{li2022coda}.

Many open-world perception methods~\cite{gupta2022owdetr, yao2024detclipv3} are proposed to address this issue.
Open-world perception tries to give precise results in dynamic and unpredictable environments, which contain novel objects and involve scene domain shifting.
Current open-world perception methods can be roughly divided into two categories: \textit{open-set} and \textit{open-ended}.
Open-set methods~\cite{zhang2022glipv2,wang2023detecting,chen2023open} often calculate the similarity between image regions and category names with a pretrained CLIP~\cite{clip} model.
Thus, they require predefined object categories as inputs for the CLIP text encoder during inference.
However, in many real-world application scenarios, we do not have the exact predefined object categories.
% For instance, in autonomous driving, self-driving vehicles may meet unexpected objects, including accident cars with fire.
For instance, in autonomous driving, self-driving vehicles may meet unexpected objects, including various rare animals.
Besides, some objects cannot be presented by a simple category name, such as a human in an animal costume, which may look like an animal but is actually a human. Some methods use generic obstacle detection to handle unknown objects. However, many things do not have a significant 3D shape, like pits or grains on the ground. Thus, open-set methods cannot handle all situations.
In contrast, open-ended methods~\cite{lin2024generateu,yao2024detclipv3} are more general and practical since they can predict the object categories and locations themselves.

In a separate line of research, large vision-language models (VLMs)~\cite{liu2024llava, li2023blip,zhu2023minigpt} show a strong generalized ability to recognize objects, \textit{e.g.}, it can recognize rare objects for corner cases in autonomous driving scenarios~\cite{wen2023gpt4v}. 
However, VLM's localization ability is less accurate than that of specific perception models~\cite{zhang2023llavaground}, sometimes missing objects or giving wrong localization results. 
On the other hand, as a pure vision model, segment-anything model (SAM)~\cite{kirillov2023sam} exhibits good generalized segmentation capabilities for images from many different domains. 
However, SAM is unable to provide categories for segmented objects~\cite{yuan2024Open-Vocabulary-SAM} and may yield numerous irrelevant segmentation results.

\begin{figure}[!t]
    \centering
    % \vspace{-10pt}
    \includegraphics[width=0.98\linewidth]{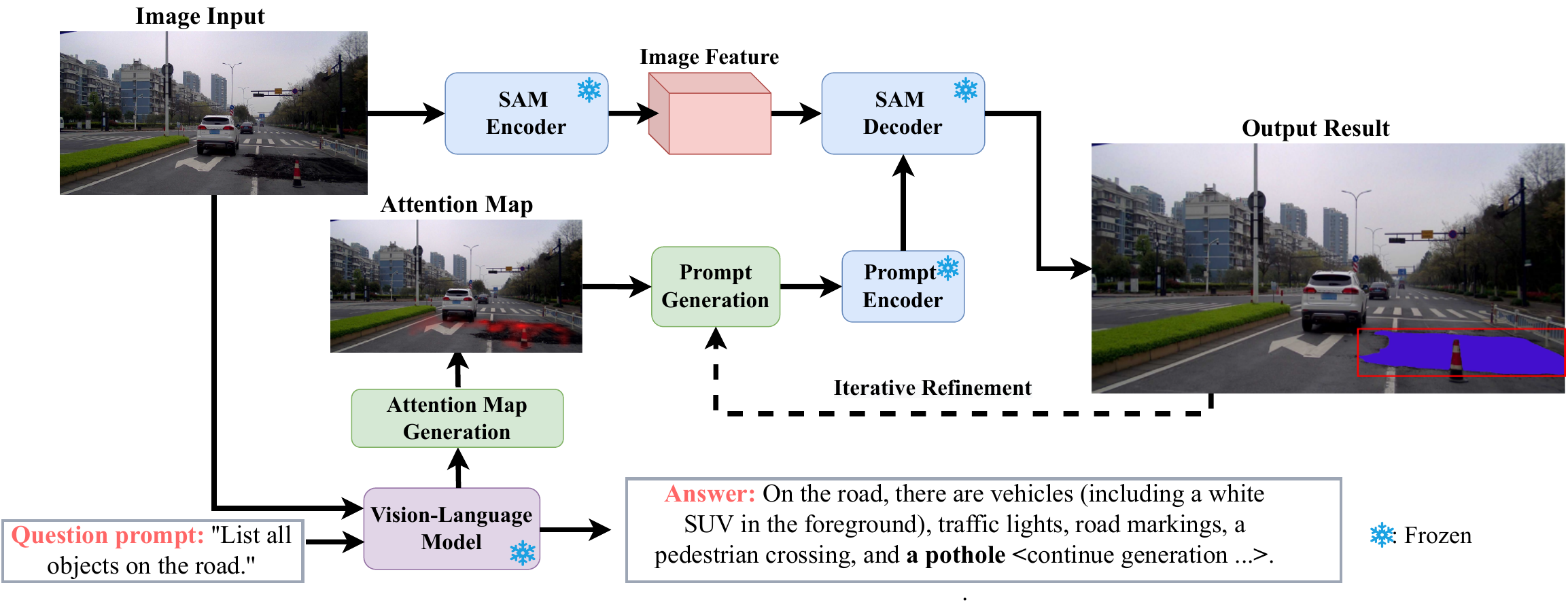}
    \vspace{-8pt}
    \caption{
    \textbf{Illustration of VL-SAM.} Without additional training, we connect the vision-language and segment-anything models with attention maps as the intermediate prompts.
    % Comparison of radar BEV features before and after alignment.
    }
    \vspace{-12pt}
    \label{fig:illustration}
\end{figure}

% In this paper, we propose to combine the existing generalized object recognition model, \textit{i.e.}, VLM, with the generalized object localization model, \textit{i.e.}, SAM, to address the object detection task in corner cases. 
In this paper, we propose to combine the existing generalized object recognition model, \textit{i.e.}, VLM, with the generalized object localization model, \textit{i.e.}, SAM, to address the open-ended object detection and segmentation task. 
We present VL-SAM, a training-free framework that connects two generalized models with attention maps as the intermediate prompts, as illustrated in Figure~\ref{fig:illustration}.
Specifically, we utilize the attention maps generated by VLM when describing the whole driving scene to prompt the segmentation of SAM. 
Firstly, given the generated token of VLM, we use the token as the query to obtain the attention maps from all layers and heads of VLM. 
Then, in the attention map generation module, we introduce the head aggregation and attention flow mechanism to aggregate and propagate global attention maps through all heads and layers. 
Besides, to alleviate the collapse problem caused by causal masks when propagating with attention flow, we adopt a regularization term to constrain the attention flow propagation process. 
After that, to better guide SAM to segment with the attention maps, we present a prompt generation module by grouping and sampling positive and negative points as the point prompts for SAM. 
Furthermore, to reduce the number of missing objects, we further use the segmentation results from SAM to sample positive and negative points from attention maps iteratively until convergence.

The main contributions of this work are summarized as follows:
% \vspace{-3pt}
\begin{itemize}
% \begin{compactitem}
    \item 
    We present VL-SAM, a training-free open-ended object detection and segmentation framework that connects the generalized object recognition model and the generalized object localization model with attention maps as the prompts.
    % , to address the object detection task for corner cases.
    \item 
   We introduce a head aggregation and regularized attention flow mechanism to aggregate and propagate attention maps with the causal masks through all heads and layers.
    \item
    We propose an iterative refinement pipeline with a positive and negative point sampling strategy for attention maps.
    \item 
    % VL-SAM outperforms current open-vocabulary detection models and achieves new state-of-the-art corner case object detection results on the CODA~\cite{li2022coda}. 
    % Meanwhile, VL-SAM achieves competitive performance on the long-tail instance segmentation dataset, LVIS~\cite{gupta2019lvis}.
    VL-SAM outperforms the \textit{open-ended} method GenerateU and obtains competitive results compared with existing \textit{open-set} methods on the long-tail instance segmentation dataset LVIS~\cite{gupta2019lvis}.
    In autonomous driving applications, VL-SAM achieves favorable corner case object detection performance on the CODA~\cite{li2022coda}. 
\end{itemize}
% \end{compactitem}

% \vspace{-3pt}
\section{Related work}
% \vspace{-3pt}
\subsection{Vision Language Model}
Large language models (LLMs), including GPT-3~\cite{gpt3}, GLM~\cite{GLM}, and LLaMA~\cite{touvron2023llama}, have shown human-like dialogue and reasoning skills. However, the limitation of LLM’s ability to process and understand visual data restricts its application to more real scenarios. To overcome this, a cutting-edge Vision-Language Model (VLM) is introduced to open up new vistas for application. 
Recently, BLIP-2~\cite{li2023blip} proposes Q-Former to connect and fuse image and text embeddings with three alignment pretrain losses. LLaMA-Adapter~\cite{zhang2023llamaadapter,gao2023llamaadapterv2}, LLaVA~\cite{liu2024llava}, and MiniGPT~\cite{zhu2023minigpt} introduce an adapter or projection layer to align the embedding space from image and text. CogVLM~\cite{wang2023cogvlm} presents visual expert modules to transform the image features to align with text features in different transformer heads. SPHINX~\cite{lin2023sphinx} utilizes several mixing techniques for multiple visual tasks. Furthermore, CogAgent~\cite{hong2023cogagent} and LLaVA-Phi~\cite{zhu2024llavaphi} view VLM as an agent or assistant to complete various tasks. 

Existing VLMs, especially GPT-4V~\cite{achiam2023gpt}, exhibit strong generalization capability for understanding and reasoning new or rare situations, \textit{e.g.,} it can deal with corner cases for autonomous driving~\cite{wen2023gpt4v}. However, the localization ability of VLMs is weaker than specific perception models, like SAM.

In this paper, we equip VLM with generalized segmentation models, \textit{i.e.}, SAM, to address the localization limitation of VLM for open-ended object detection and segmentation. We achieve this by connecting two models with attention maps as the prompts without additional training.

\subsection{Open-World Object Detection and Segmentation}
% \justificationTODO{}
% With the advent of the CLIP models~\cite{clip}, open-world classification has made great progress, and the development of open-world object detection has been promoted at the same time.
With the advent of the CLIP models~\cite{clip}, open-world classification, object detection, and instance segmentation have made great progress at the same time.
Open-world methods try to discover and recognize unseen objects in the training set during inference.
Current open-world methods can be roughly classified into two types: \textit{open-set}~\cite{scheirer2012openset} and \textit{open-ended}~\cite{lin2024generateu}.
Open-set methods require redefined object categories, including seen objects and unseen objects in the training set, as inputs during inference.
By contrast, open-ended methods can locate seen and unseen objects and generate their names simultaneously, as the current VLM does.
In real-world applications, the exact categories may remain unknown for the perception models. 
For instance, in autonomous driving, self-driving vehicles often encounter unknown objects on the road, including overturned cars and construction vehicles with various shapes.
Thus, the open-ended problem is more general and practical.

\noindent\textbf{Open-Set Methods.} 
With the powerful text-image embedding matching with CLIP, current open-set object detection methods mainly use a proposal network to obtain foreground object bounding boxes and embeddings, and then use CLIP as the open-set classification module to predict their categories.
More recently, GLIP~\cite{glip} proposes to use phrase grounding to pre-train open-world object detectors. 
GroundingDINO~\cite{liu2023groundingdino} presents cross-modality fusions to introduce text information to the image encoder for object grounding. 
SWORD~\cite{wu2023exploring} designs a novel contrastive method to learn the discrimination between foreground and background for instance segmentation. 
YOLO-World~\cite{cheng2024yoloworld} introduces a prompt-then-detect paradigm for real-time open-world object detection. 
However, the above methods require predefined object categories as inputs for the text encoder. 

\noindent\textbf{Open-Ended Methods.}
GenerateU~\cite{lin2024generateu} first proposes the open-ended problem.
Concurrently, DetCLIPv3~\cite{yao2024detclipv3} introduces a similar concept with open-ended.
They present a generative framework with language models to generate object categories and bounding boxes at the same time.
To achieve better generalization capabilities, they construct a large dataset with bounding box and caption pairs and finetune the whole network on the constructed dataset.

In contrast, we propose a training-free open-ended framework, VL-SAM, that combines generalized recognition and segmentation models. VL-SAM can generate object categories with the generalized recognized model and then localize objects with the generalized segmentation models.

\begin{figure}[!t]
    \centering
    % \vspace{-10pt}
    \includegraphics[width=1.0\linewidth]{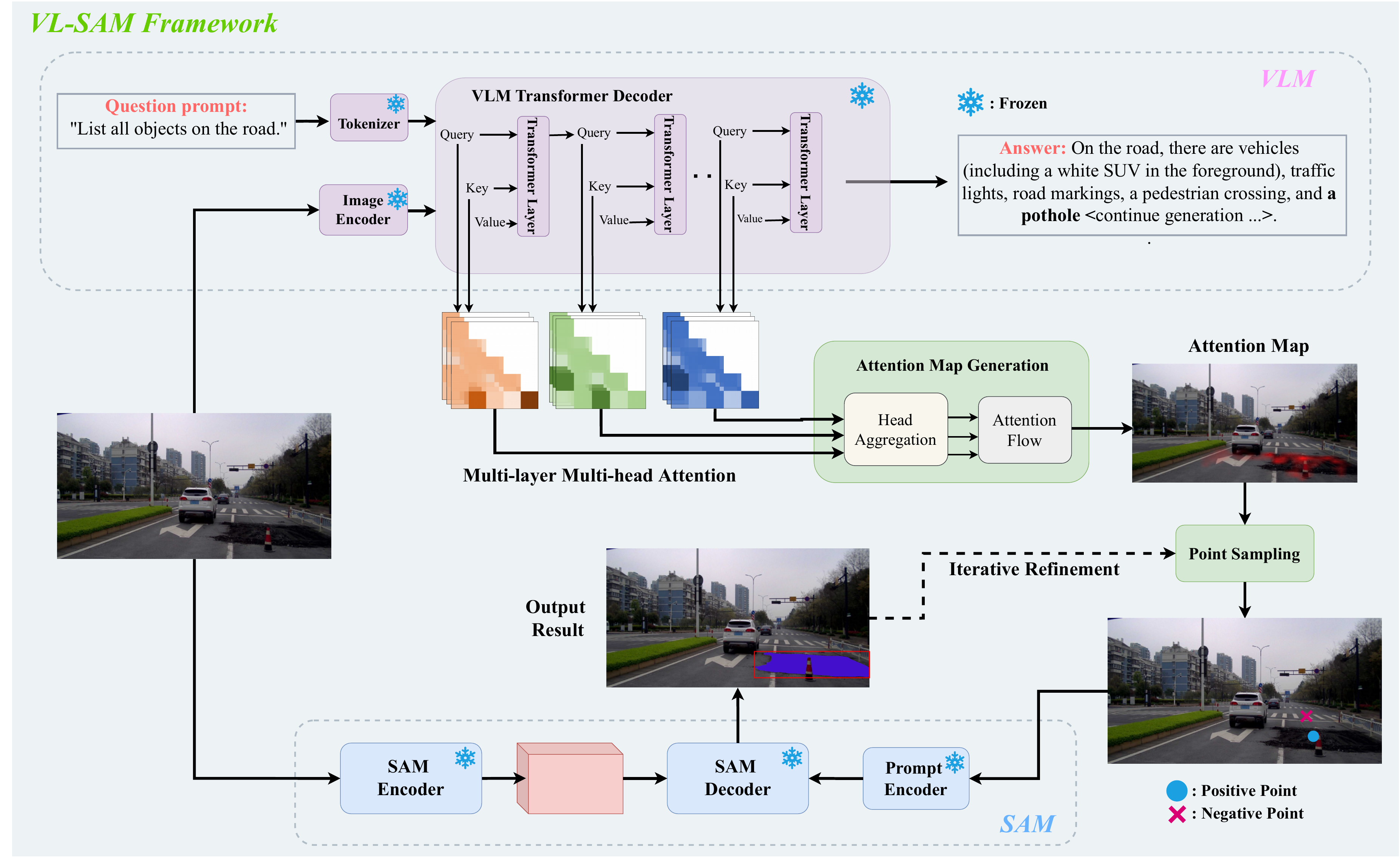}
    % \vspace{-10pt}
    \caption{
    \textbf{An overview of VL-SAM framework.} We first use VLM to describe the input image and generate all possible objects' names. Then, for each object name, we obtain the corresponding attention map with the attention map generation module. Finally, we sample point prompts from the attention map and send them to SAM to predict detection and segmentation results.}
    \label{fig:pipeline}
\end{figure}

\section{Method}
As shown \ref{fig:pipeline}, we provide an overview of our proposed framework. We use VLM and SAM as the generalized object recognition model and object localization model, respectively. Given an image input, we first use VLM to describe the scene and list all possible objects in the image. Then, for each object, we use the attention generation module with head aggregation and attention flow to obtain the high-quality attention map from VLM. Finally, we generate point prompts from the attention map and send them to SAM to get the location prediction iteratively.

\subsection{Preliminary}

\noindent \textbf{Segment Anything Model.}
SAM is a prompt-based segmentation model with excellent data generation capability. It consists of three components: an image encoder, a mask decoder, and a prompt encoder. SAM takes an image and a set of prompts, including points, a box, and a mask, as the inputs. To segment objects with the prompts, SAM first extracts image features with the image encoder. Concurrently, the set of prompts is sent to the prompt encoder to transform into the prompt tokens. Then, the image features, prompt tokens, and mask tokens interact in the mask decoder with the two-way transformers. Finally, the mask tokens are transformed into multi-scale segmentation masks by multiplying mask tokens with the image features following MaskDINO~\cite{li2023maskdino}. 

\noindent \textbf{Auto-Regressive Based Vision-Language Model.}
Current Auto-Regressive based VLMs have yielded surprising performance in various vision-language tasks. The mainstream framework of current VLMs comprises four parts, \textit{i.e.}, an image encoder, a text tokenizer, projection layers, and a language decoder. Given an image and text as inputs, VLMs extract image tokens and text tokens with the image encoder and text tokenizer, respectively. Then, the image tokens are aligned with text tokens with projection layers. After that, the tokens from two modals are concatenated and sent to the language decoder to generate text outputs. The language decoder adopts the next-token prediction paradigm that the probability of the current generated token $x_t$ depends on all previous tokens $(x_1, x_2, ..., x_{t-1})$.

\begin{figure}[!t]
    \centering
    % \vspace{-10pt}
    \includegraphics[width=0.6\linewidth]{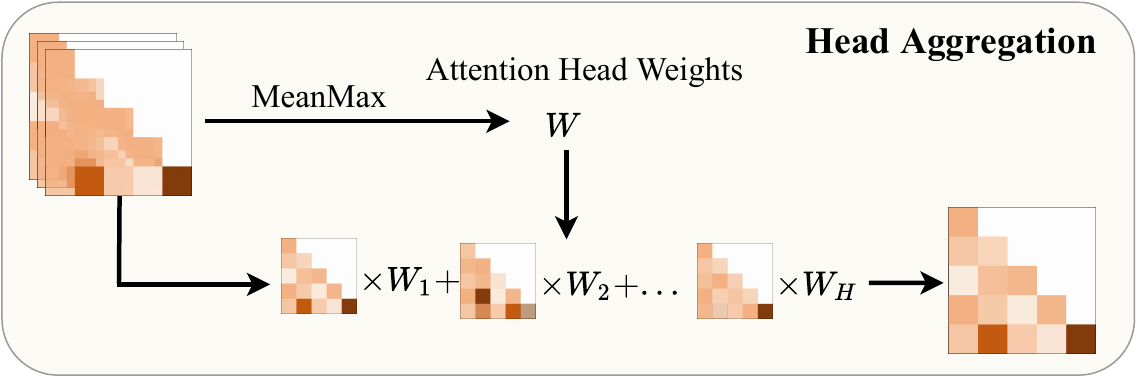}
    % \vspace{-10pt}
    \caption{
    \textbf{Head aggregation.} We aggregate information from all attention heads with head weights.}
    \vspace{-12pt}
    \label{fig:head}
\end{figure}

\subsection{Attention Map Generation}
\label{sec:Attention Map Generation}
The main idea of VL-SAM is to use attention maps of objects as the prompts for SAM to segment. Thus, how to generate a high-quality attention map for an object is critical. To achieve this, we introduce attention flow to aggregate and propagate attention maps through all transformer heads and layers in VLM.

Specifically, given an image input, we ask VLM to give all possible objects in the image. During this process, we cache all queries and keys from VLM. Then, we multiply queries and keys with causal masks and SoftMax normalization to obtain similarity matrix $S\in N\times N\times H\times L$, where $N$ is the length of queries and keys, $H$ is the number of transformer heads, and $L$ denotes the number of transformer layers. $S_{i,j}^{h,l}$ represents the similarity between query $i$ and key $j$ in the head $h$, layer $l$. After that, we aggregate information from all transformer heads with mean-max attention head weights, as shown in Figure~\ref{fig:head}. In particular, we choose the maximum similarity weights of matrix $S$ in dimension $j$ and average them in dimension $i$ to obtain the attention head weights $W\in 1\times 1\times H\times L$:
\begin{equation}
    W = \texttt{Mean}(\texttt{Max}(S, ~dim=1), ~dim=0).
\end{equation}
Obviously, the attention head weight indicates the importance of each head in each layer. Then, we pointwise multiply attention head weight $W$ with similarity matrix $S$ and average all heads as follows:
\begin{equation}
    S' = \texttt{Mean}(S\odot W,~ dim=2).
\end{equation}

\begin{figure}[!t]
    \centering
    % \vspace{-10pt}
    \includegraphics[width=0.8\linewidth]{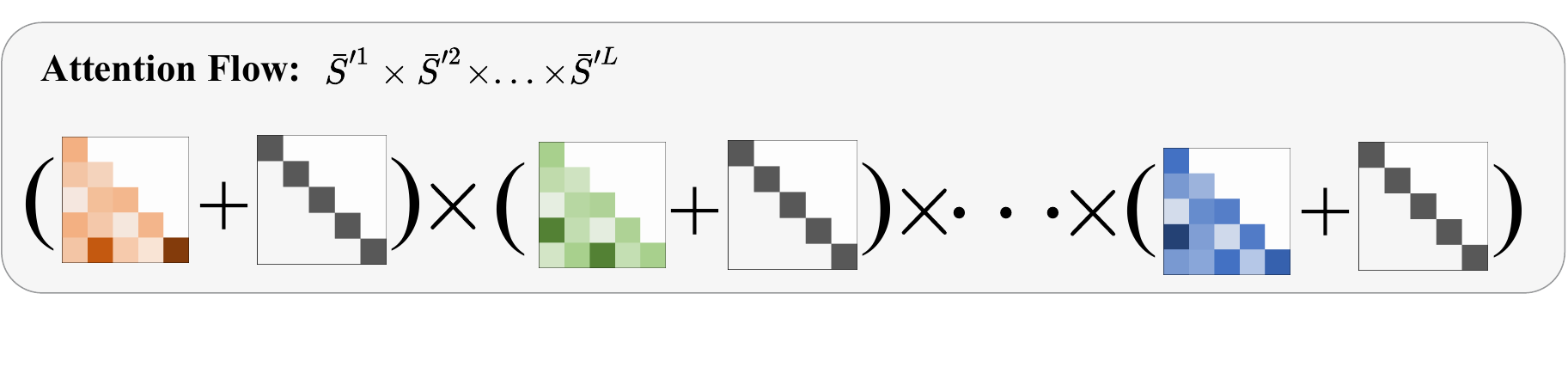}
    \vspace{-20pt}
    \caption{
    \textbf{Attention flow.} We propagate attention from the first layer to last layer with attention flow.}
    \label{fig:attn}
\end{figure}

\begin{figure}[!t]
    \centering
    % \vspace{-10pt}
    \includegraphics[width=1.0\linewidth]{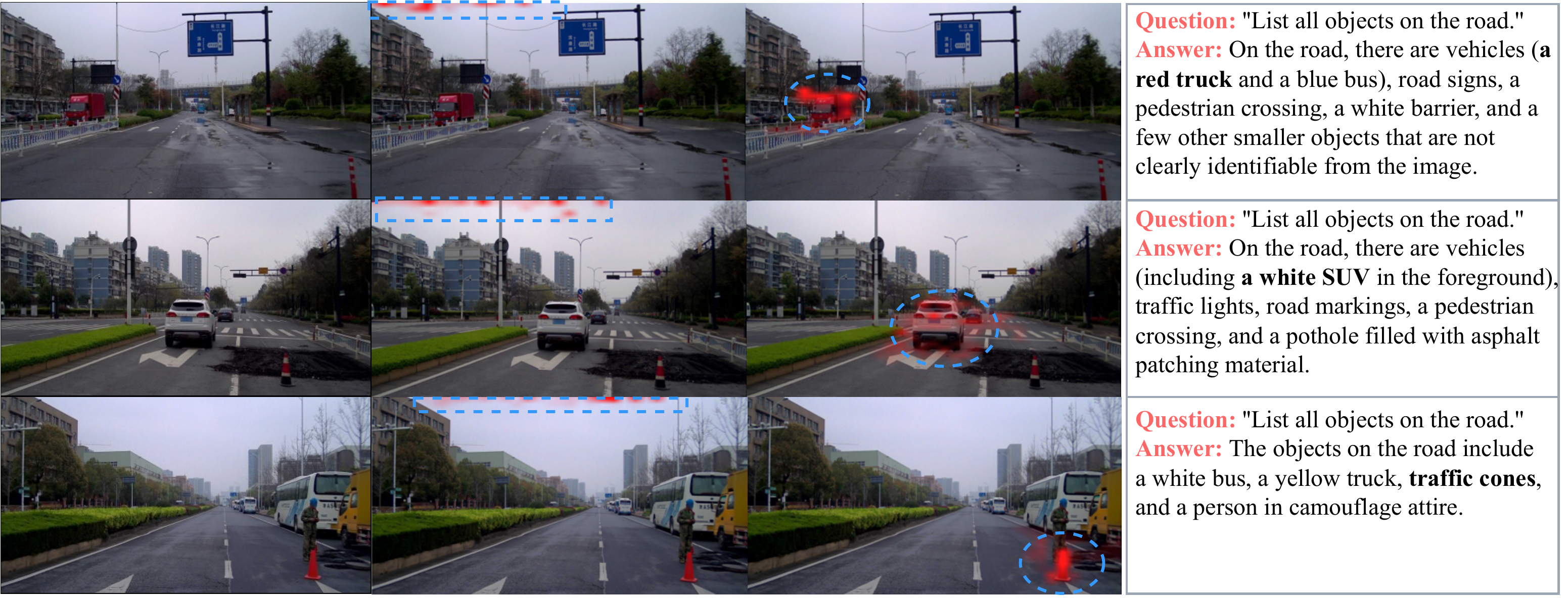}
    % \vspace{-30pt}
    \caption{
    \textbf{Illustration of attention collapse.} For each column, from left to right, we show image inputs, attention flow (collapse), regularized attention flow, and generated answers from VLM.}
    \label{fig:degradation}
\end{figure}

After aggregating all information from all heads, we present attention flow to further aggregate attention from all layers, as illustrated in Figure~\ref{fig:attn}. Concretely, we use the attention rollout method~\cite{attentionflow} to compute the attentions from layer $l-1$ to layer $l$ as follows:
\begin{equation}
    \bar{S'}_{i,j}^{l} = \sum_{k=1}^{N}(I_{i,k}+S_{i,k}^{'l})\times(I_{k,j}+\bar{S'}_{k,j}^{l-1}),
\end{equation}
where $I$ is the identity matrix.
After the attention rollout, we only need the attention map from the last layer.
% To obtain the similarity between the generated token (object) and image feature, we select the corresponding line and columns.
To obtain the image attention map of the generated token, we select the corresponding line and columns from $\bar{S'}^{L}$.

However, since VLM uses causal masks for auto-regressive generation, simply adopting the attention rollout method causes attention collapse, as shown in Figure~\ref{fig:degradation}. 
% To alleviate this problem, we introduce regularized attention rollout.
Fortunately, we find a simple regularization term that can alleviate this problem efficiently.
Specifically, for each column, assuming the unmasked length is $L_0$, we multiply each value in this column with $1-(L_0-1)/L$. 
%
% With this regularization term, the value in the top left corner will be relatively smaller, and the value in the bottom right corner will be larger. Thus, the value balance in the attention map is kept.
With this regularization term, the attention value in the top left corner will be constrained.

\subsection{SAM Prompt Generation}
\label{sec:Prompt Generation}
The attention map generated in Section~\ref{sec:Attention Map Generation} has some unstable false positive peaks. To filter these false positive areas, we first use a threshold to filter weak activated areas and find the maximum connectivity area as the positive area~\cite{caron2021dino}. The remaining area serves as a negative area.
After that, we sample a positive point from the positive area with the maximum activated value and a negative point from the negative area with the weakest activated value. The positive and negative points serve as the point prompt pair for SAM.

\subsection{Iterative Refinement}
The segmentation results from the SAM decoder may include rough edges and background noises. We adopt two iterative strategies to further refine the segmentation results.
In the first iterative strategy, we follow the cascaded post-refinement in PerSAM~\cite{zhang2023persam} to take the initial segmentation masks generated with the positive and negative pairs as the additional prompt input for the SAM decoder. For the second iterative strategy, we use the segmentation masks in the first iterative strategy to mask the attention map $\bar{S}'$. Then, we iteratively generate positive and negative pairs with Prompt Generation in Section~\ref{sec:Prompt Generation} from the masked attention map and send them to the SAM decoder. Finally, we aggregate the results with NMS~\cite{girshick2015fastrcnn}.
% and select segmentation masks with the Top-k IoU score output by the SAM decoder.

\subsection{Multi-scale Ensemble}
Due to the low-resolution image input of the image encoder in VLM, VLM may fail to recognize small objects. For instance, it may generate an answer: `\textit{On the road, there are vehicles (a red truck and a blue bus), road signs, a pedestrian crossing, a white barrier, and \textbf{a few other smaller objects that are not clearly identifiable from the image.}}'. To alleviate this issue, we follow SPHINX~\cite{lin2023sphinx} to split an image ($H\times W$) into four sub-images ($H/2\times W/2$) from the four corners and send each sub-image to VL-SAM independently. Finally, we ensemble the output of VL-SAM for four sub-images and the whole image.

\subsection{Question-prompt Ensemble}
The output of VLM is sensitive to the input prompt. To obtain a more comprehensive description of the input image, we ask VLM to generate ten question prompts for scene description with the sentence:`\textit{If we want you to list all possible objects in the given image, what questions should we ask? Please give 10 questions you prefer.}' Then, we use the generated question prompts for VL-SAM to segment objects and ensemble the outputs of all question prompts.

\section{Experiments}

\subsection{Implementation Details}
\label{Implementation Details}
We chose CogVLM-17B~\cite{wang2023cogvlm} with EVA2-CLIP-E~\cite{sun2023eva} and Vicuna-7B-v1.5~\cite{chiang2023vicuna} as the vision-language model.
CogVLM-17B divides an image with $490\times 490$ into $35\times 35$ patches. We set the temperature to 0.8 and top-p for nucleus sampling to 0.1 for CogVLM-17B.
For the generated localization model, we use SAM with ViT-Huge~\cite{dosovitskiy2020vit}.

We evaluate VL-SAM in a \textit{training-free zero-shot} manner for all datasets.
To obtain object categories from the generated sentence of VLM, we follow Tag2Text~\cite{huang2023tag2text} to parse tags from the given sentence.
To evaluate the open-ended performance on datasets with predefined object category names, we follow GenerateU~\cite{lin2024generateu} to adopt CLIP~\cite{clip} text encoder and map the generated object categories to predefined categories in datasets for evaluation. Specifically, we use the text prompt `a \{object category\}' for CLIP text encoder to calculate the similarity between generated object categories and predefined categories for mapping.
% In addition to our method, we implement and test two naive open-ended baselines that incorporate VLM. For both baselines, we use VLM to generate the object categories in the same manner as VL-SAM with question ensemble techniques. Then, we adopt GroundingDINO to ground corresponding objects with the object categories as the first baseline.
% For the second baseline, we utilize an open-world detection model, YOLOWorld, to localize objects with the object categories as the classification branch inputs.
All models are inferred on an 80G A800 machine.

% \subsection{Datasets}

\begin{table}[t]
  \caption{\textbf{Comparison of object detection and segmentation results on LVIS minival.}
  `Open-Ended' denotes that we do not have exact object categories during inference~\cite{lin2024generateu}. We report \textit{fixed} AP~\cite{dave2021fixedap} for rare objects.
  * denotes using the external data.
  }
  \label{tab:lvis}
  \centering
  \begin{tabular}{l|c|c|cc}
    \toprule
    \multirow{2}{*}{Method} & \multirow{2}{*}{Type}& \multirow{2}{*}{Training} & \multicolumn{2}{c}{LVIS}\\
    &&& box AP$_{rare}$ & mask AP$_{rare}$ \\
    \midrule
    Mask R-CNN~\cite{he2017maskrcnn} &\multirow{2}{*}{Close-Set} & $\checkmark$& 26.3 & 25.1\\
    Deformable DETR~\cite{zhu2020deformable} & &$\checkmark$ & 24.2 & - \\
    \midrule
    GLIP~\cite{glip}&\multirow{5}{*}{Open-Set}& $\checkmark$&20.8 & - \\
    % GLIPv2&& - & - & - & 29.0 \\
    GroundingDINO~\cite{liu2023groundingdino}&& $\checkmark$&27.4 & - \\
    DetCLIP~\cite{yao2022detclip} &&$\checkmark$&26.9 & -\\
    YOLOWorld~\cite{cheng2024yoloworld} && $\checkmark$ & 27.1 & -\\
    OWLv2$^{*}$~\cite{owlv2} && $\checkmark$ & 39.0 & -\\
    \midrule
    % GenerateU &\multirow{3}{*}{Open-Ended} &5.3  \\
    % GenerateU (Full-Finetune) & &20.0\\
    GenerateU~\cite{lin2024generateu} &\multirow{2}{*}{Open-Ended} & $\checkmark$&20.0 & - \\
    VL-SAM (Ours)& & $\times$ & 23.4 & 22.7 \\
    \bottomrule
  \end{tabular}
\end{table}

\subsection{Main Results}
% here
\noindent \textbf{LVIS Dataset.} 
% To further demonstrate the effectiveness of the proposed method for rare object discovery, we evaluate VL-SAM on the LVIS dataset~\cite{gupta2019lvis}, which has a long tail of categories and annotations for over 1000 object categories.
We evaluate VL-SAM on the LVIS dataset~\cite{gupta2019lvis}, which has a long tail of categories and annotations for over 1000 object categories.
Following previous works~\cite{lin2024generateu, cheng2024yoloworld}, we mainly evaluate VL-SAM on LVIS minival and report the fixed AP~\cite{dave2021fixedap} for rare objects.

As shown in Table~\ref{tab:lvis}, we list the performance for three types of perception methods, \textit{i.e.,} close-set, open-set~\cite{gupta2022owdetr}, and open-ended.
The different between open-set and open-ended is that open-set requires exact prior knowledge of object categories as inputs, while open-ended can generate them during inference in a zero-shot manner~\cite{lin2024generateu}.
In a real scenario, we often do not have predefined object categories for a scene. Thus, open-ended methods are more general and practical.
% \justificationTODO{}
As we can see, VL-SAM outperforms GenerateU by 3.4 $AP_{rare}$.
Notably, VL-SAM is a training-free framework and can simultaneously obtain boxes and segmentation masks. In contrast, GenerateU needs to fine-tune both the image encoder and language model on VG~\cite{krishna2017vg} and GRIT~\cite{peng2023instructiongrit} datasets, requiring significant training costs, and can only predict bounding boxes.
Besides, VL-SAM achieves competitive detection and segmentation performance compared to open-set detection methods and close-set segmentation methods, respectively.

\begin{table}[t]
  \caption{\textbf{Comparison of object detection results on CODA.} We chose the best performance for $^*$ results from CODA. $^{\dagger}$ denotes few-shot object detectors in the one-shot setting. `Oracle' represents utilizing ground-truth boxes as the box prompt for SAM.}
  \label{tab:coda}
  \centering
  \begin{tabular}{l|c|c|c|ccc}
    \toprule
    % \multicolumn{2}{c}{Part}                   \\
    % \cmidrule(r){1-2}
    \multirow{2}{*}{Method} & \multirow{2}{*}{Type} & \multirow{2}{*}{VLM} & \multirow{2}{*}{Training} & \multicolumn{3}{c}{CODA}\\
    % \cmidrule(r){3-5}
    &&&& mAR & AR$_{50}$ & AR$_{75}$\\
    \midrule
    % CODA$^*$ & 10.6 & 20.0 & 10.2     \\
    % RPN (Cascade) & Output terminal & $\sim$10      \\
    RetinaNet$^*$~\cite{retinanet}& \multirow{7}{*}{Close-Set} & $\times$ & $\checkmark$ & 12.8 &23.2 &11.9\\
    Faster R-CNN$^*$~\cite{ren2015fasterrcnn}& & $\times$ & $\checkmark$& 10.7& 19.2& 10.2\\
    Cascade R-CNN$^*$~\cite{cai2018cascadercnn}& &$\times$ & $\checkmark$& 10.4 &18.5& 9.7 \\
    Deformable DETR$^*$~\cite{zhu2020deformable}&& $\times$ & $\checkmark$ & 9.0 &22.2& 5.6 \\
    Sparse R-CNN$^*$~\cite{sun2021sparsercnn}&& $\times$ & $\checkmark$ & 10.1 &19.6 &9.0\\
    Cascade Swin$^*$~\cite{liu2021swin}&& $\times$ & $\checkmark$ & 9.9 &17.2& 9.7 \\
    RPN$^*$~\cite{ren2015fasterrcnn}&& $\times$ & $\checkmark$ & 10.6 &20.0 &10.2 \\
    \midrule
    ORE$^*$~\cite{joseph2021ore}& \multirow{5}{*}{Open-Set} & $\times$ & $\checkmark$     & 8.3 &16.4 &7.4  \\
    FsDet$^{\dagger}$~\cite{FSDet}&& $\times$ & $\checkmark$ & 4.2& 7.7& 4.0 \\
    DeFRCN$^{\dagger}$~\cite{DeFRCN}& & $\times$ & $\checkmark$ & 4.5& 8.9& 4.2 \\
    % \midrule
    % CogVLM~\cite{wang2023cogvlm}+GroundingDINO~\cite{liu2023groundingdino}&$\checkmark$ & 12.6 & 21.7 & 13.3 \\
    % CogVLM~\cite{wang2023cogvlm}+YOLOWorld~\cite{cheng2024yoloworld}&$\checkmark$ &16.1 & 26.2 & 19.6\\
    GroundingDINO~\cite{liu2023groundingdino}& & $\checkmark$ & $\checkmark$ & 12.6 & 21.7 & 13.3 \\
    YOLOWorld~\cite{cheng2024yoloworld}& & $\checkmark$ & $\checkmark$ &16.1 & 26.2 & 19.6\\
    \midrule
    LLaVA-Grounding~\cite{zhang2023llavaground}& \multirow{2}{*}{Open-Ended} &$\checkmark$ & $\checkmark$ & 18.4 & 30.5 & 22.0\\
    % GenerateU~\cite{lin2024generateu} & $\checkmark$ & 33.7 \\
    % \midrule
    VL-SAM (Ours)& & $\checkmark$ & $\times$ & 40.1 & 90.1 & 50.5\\
    % VL-SAM-ensemble (Ours)&$\checkmark$ & 48.9 & 92.8 & 59.2 \\
    \midrule
    GT+SAM (Oracle)&$-$ &$-$ &$-$ & 54.1 & 94.1 & 64.9\\

    \bottomrule
  \end{tabular}
\end{table}

\noindent \textbf{CODA Dataset.} 
To further demonstrate the effectiveness of the proposed method in the real-world application, we present the results of VL-SAM on corner case object detection dataset CODA for autonomous driving in Table \ref{tab:coda}. 
Specifically, as we can see, RPN only achieves 10.6 mAR, indicating that current open-set detectors relying on object proposals have difficulty dealing with corner cases.
For more recent open-set detectors, they achieve higher mAR with CLIP as the object category predictor.
% Besides, we find that ensembling VLM and grounding models into one model, \textit{e.g.,} LLaVA-Grounding, achieves better performance than simple cascading VLM and grounding or open-world models for corner cases.
% However, aggregating VLM and grounding models to one model generally requires joint training of two models, introducing additional training costs.
%
For the open-ended method, LLaVA-Grounding ensembles VLM and grounding models into one model and achieves better performance than open-set methods.
However, aggregating VLM and grounding models to one model requires joint training of two models, introducing additional training costs.
By contrast, VL-SAM is a training-free framework and obtains significant performance improvement over LLaVA-Grounding from 18.4 mAR to 40.1 mAR. 
% Specifically, current open-world detection methods with VLM surpass previous methods.

In addition, we evaluate the performance upper bound of the current SAM. We utilize ground-truth boxes as the box prompt for SAM decoder to segment objects. We can observe that, in this setting, SAM achieves 54.1 mAR and 94.1 AR$_{50}$ since SAM has its limitations on segmentation tasks. It sometimes over- or under-segments an object and cannot obtain perfect segmentation results. 
% Thus, it cannot obtain
% Second, some ground-truth boxes of CODA are oversized since some boxes are the boundary of projected 3D boxes to the 2D image plane, as shown in \ref{fig:coda}.
%
Nevertheless, VL-SAM achieves 74.1\% mAR performance of this upper bound, demonstrating the effectiveness of the proposed framework.
Overall, VL-SAM achieves favorable performance on the CODA dataset.

\subsection{Ablation Study}

\textbf{Main Components.} As shown in Table \ref{tab:ablation}, we conduct ablation studies on CODA to analyze the effectiveness of each component of VL-SAM.
For the baseline Naive Attention method, we use the attention map from the last layers and average all attention heads.
We can see that the Naive Attention baseline obtains unsatisfactory results even with multi-scale and question ensemble techniques.
With the proposed attention generation module, we improve the baseline by 7.9 mAR.
Adding points pairs with prompt generation brings 2.2 mAR improvement.
Besides, refining the segmentation maps with the iterative refinement module improves the detection performance from 12.3 mAR to 14.1 mAR.
Furthermore, ensembling with multi-scale image input and question prompt obtains 13.2 mAR and 12.8 mAR, respectively.
Though multi-scale and question prompt ensembles greatly improve performance, these two ensemble techniques do not show effectiveness without the proposed components.
In summary, the results show the effectiveness of each component proposed in VL-SAM.

\begin{table}[t]
  \caption{\textbf{Ablation of main components.} `Attn' is short for `attention'. Each component improves the detection performance consistently.}
  \label{tab:ablation}
  \centering
  \resizebox{0.98\textwidth}{!}{
  \begin{tabular}{c|ccc|cc|ccc}
    \toprule
    % \multicolumn{2}{c}{Part}                   \\
    % \cmidrule(r){1-2}
    % \multirow{2}{*}{Method} & \multirow{2}{*}{VLM}    & \multicolumn{3}{c}{CODA}\\
    % \cmidrule(r){3-5}
    Naive Attn &Attn Generation & Prompt Generation & Iterative Refine & Multi-scale & Question ensemble & mAR\\
    \midrule
    % $\checkmark$ & 12.6 & 21.7 & 13.3 \\
    $\checkmark$ & &  & &  & & 2.2 \\
    $\checkmark$ & &  & & $\checkmark$ & $\checkmark$ & 5.0 \\
    \midrule
    & $\checkmark$ &  & &  & & 10.1 \\
    & $\checkmark$ & $\checkmark$ & &  & & 12.3 \\
    & $\checkmark$ & $\checkmark$ & $\checkmark$ &  & & 14.1 \\
    & $\checkmark$ & $\checkmark$ & $\checkmark$ & $\checkmark$ & & 27.3 \\
    & $\checkmark$ & $\checkmark$ & $\checkmark$ & $\checkmark$ & $\checkmark$ & 40.1 \\
    \bottomrule
  \end{tabular}
  \vspace{-15pt}
  }
\end{table}

\textbf{Attention Generation.}
To obtain high-quality attention maps from VLM, we introduce head weights to fuse transformer heads and a regularization term for attention flow. 
As shown in Table \ref{tab:attn}, simply using attention flow~\cite{attentionflow} almost fails to recognize objects for SAM due to the attention collapse caused by causal masks (Figure~\ref{fig:degradation}).
With the regularization term, the attention flow mechanism shows its superiority over naive attention by improving 6.3 mAR.
% fusing with head weights leads to xxx mAR improvement. 
%
Moreover, fusing with head weights leads to a 1.6 mAR improvement.

\begin{table}[t]
  \caption{\textbf{Ablation of attention generation.} We can obtain high-quality attention maps with the proposed modules. }
  \label{tab:attn}
  \centering
  % \resizebox{0.98\textwidth}{!}{
  \begin{tabular}{c|cc|c|c}
    \toprule
    % \multicolumn{2}{c}{Part}                   \\
    % \cmidrule(r){1-2}
    % \multirow{2}{*}{Method} & \multirow{2}{*}{VLM}    & \multicolumn{3}{c}{CODA}\\
    % \cmidrule(r){3-5}
    \multirow{2}{*}{Naive Attention Map} & \multicolumn{2}{c|}{Attention Flow} &\multirow{2}{*}{Head Weight} &  \multirow{2}{*}{mAR}\\
    
     &No Regularization& Regularization & &\\
    \midrule
    % $\checkmark$ & 12.6 & 21.7 & 13.3 \\
    $\checkmark$& & &  & 2.2 \\
    \midrule
    & $\checkmark$ &  &  & 0.1 \\
    % \midrule
    & & $\checkmark$ &  &  8.5\\
     & & $\checkmark$& $\checkmark$ &  10.1\\
    \bottomrule
  \end{tabular}
  % }
\end{table}

\textbf{Model Generalization.}
% To demonstrate that VL-SAM is 
To demonstrate the model generalization ability of the VL-SAM framework, we adapt two additional popular VLMs, MiniGPT-4~\cite{zhu2023minigpt} and LLaVA~\cite{liu2024llava} to replace CogVLM and use MobileSAM~\cite{mobile_sam} to replace SAM.
In Table ~\ref{tab:generalization}, we present the results of using these models in the VL-SAM framework.
Empirical results show that replacing CogVLM with MiniGPT-4 or LLaVA may reduce the object localization performance in corner cases as CogVLM shows more powerful multimodal chat and reasoning ability than MiniGPT-4 and LLaVA. This indicates that our VL-SAM framework can benefit from more powerful VLMs.
Besides, replacing SAM with a more lightweight but less accurate MobileSAM also leads to performance drops.
Nevertheless, all these results outperform previous methods (18.4 mAR) in Table ~\ref{tab:coda}.
This evidences that our framework can generalize to multiple vision-language and segmentation models.

\begin{table}[t]
  \caption{\textbf{Ablation of model generalization.} VL-SAM can adopt various vision-language models and segmentation models.}
  \label{tab:generalization}
  \centering
  % \resizebox{0.98\textwidth}{!}{
  \begin{tabular}{c|c|c}
    \toprule
    Vision-Language Model & Segmentation Model & mAR \\
    \midrule
    CogVLM & SAM & 40.1 \\
    \midrule
     MiniGPT-4 & SAM & 34.7 \\
     LLaVA & SAM & 37.2 \\
    \midrule
    CogVLM & MobileSAM & 29.2 \\
    \bottomrule
  \end{tabular}
  \vspace{-5pt}
  % }
\end{table}

\section{Limitations}
\label{limitation}
Since we combine VLM and SAM to address the open-ended object detection and segmentation task, VL-SAM inherits the defects of VLM and SAM.
The first defect is the hallucination problem in VLM. VL-SAM also suffers from hallucinations, generating wrong object tokens and attention maps.
The second defect is the low inference speed of VL-SAM. However, these defects can be fixed in the future.
For example, there are many more efficient SAM variant models, including EfficientSAM~\cite{xiong2023efficientsam} and MobileSAM~\cite{mobile_sam}. 
Our framework can benefit from these new models since we can easily replace CogVLM and SAM in VL-SAM with these more efficient and highly accurate models.

\section{Conclusion}
\label{conclusion}
In this paper, we introduce VL-SAM, a framework that cascades VLM and SAM with the attention map to address the open-ended object detection and segmentation task. 
% Without additional training, we adopt the attention-flow mechanism to aggregate attention maps from VLM as the prompts for SAM to segment objects iteratively.
Without additional training, we adopt attention maps generated by VLM as the prompts for SAM to segment objects.
We introduce the attention flow mechanism to aggregate high-quality attention maps.
Besides, we present an iterative refinement pipeline with positive and negative points pair sampling strategy to acquire more accurate segmentation masks.
%
% Experimental results on CODA show that VL-SAM beats former state-of-the-art open-vocabulary object detectors with VLM by a clear margin of 21.7 mAR.
% %
% Besides, VL-SAM outperforms the open-ended method GenerateU on the long-tail generic instance segmentation dataset LVIS by 3.4 box AP$_{rare}$.
Experimental results on the long-tail generic instance segmentation dataset LVIS show that VL-SAM beats the open-ended method GenerateU and achieves competitive performance compared with close-set and open-set methods. 
% former state-of-the-art open-vocabulary object detectors with VLM by a clear margin of 21.7 mAR.
%
Moreover, VL-SAM achieves favorable results on the corner case object detection dataset CODA.

\noindent \textbf{Broader Impacts Statement.}
This paper studies utilizing VLM and SAM for open-ended object detection and segmentation. We do not see potential privacy-related issues.
This study may inspire future research on open-ended perception and potential corner case object detection applications in autonomous driving.
However, the proposed model's performance is not yet up to the level of practical application and may pose safety threats when applied directly in practice.
% However, the proposed model's detection performance is not yet up to the level of practical application and may pose safety threats when applied directly in practice.
% For the positive impact, this study may inspire future research on open-ended perception and potential corner case object detection applications in autonomous driving.

\newpage
\small
\bibliographystyle{plain}
\bibliography{main}

\begin{thebibliography}{10}

\bibitem{attentionflow}
Samira Abnar and Willem~H. Zuidema.
\newblock Quantifying attention flow in transformers.
\newblock In Dan Jurafsky, Joyce Chai, Natalie Schluter, and Joel~R. Tetreault, editors, {\em {Annual Meeting of the Association for Computational Linguistics (ACL)}}, 2020.

\bibitem{achiam2023gpt}
Josh Achiam, Steven Adler, Sandhini Agarwal, Lama Ahmad, Ilge Akkaya, Florencia~Leoni Aleman, Diogo Almeida, Janko Altenschmidt, Sam Altman, Shyamal Anadkat, et~al.
\newblock Gpt-4 technical report.
\newblock {\em arXiv preprint arXiv:2303.08774}, 2023.

\bibitem{gpt3}
Tom Brown, Benjamin Mann, Nick Ryder, Melanie Subbiah, Jared~D Kaplan, Prafulla Dhariwal, Arvind Neelakantan, Pranav Shyam, Girish Sastry, Amanda Askell, et~al.
\newblock Language models are few-shot learners.
\newblock {\em {Neural Information Processing Systems (NeurIPS)}}, 2020.

\bibitem{cai2018cascadercnn}
Zhaowei Cai and Nuno Vasconcelos.
\newblock Cascade r-cnn: Delving into high quality object detection.
\newblock In {\em {IEEE Conference on Computer Vision and Pattern Recognition (CVPR)}}, 2018.

\bibitem{caron2021dino}
Mathilde Caron, Hugo Touvron, Ishan Misra, Herv{\'e} J{\'e}gou, Julien Mairal, Piotr Bojanowski, and Armand Joulin.
\newblock Emerging properties in self-supervised vision transformers.
\newblock In {\em {IEEE International Conference on Computer Vision (ICCV)}}, 2021.

\bibitem{chen2023open}
Xi~Chen, Shuang Li, Ser-Nam Lim, Antonio Torralba, and Hengshuang Zhao.
\newblock Open-vocabulary panoptic segmentation with embedding modulation.
\newblock In {\em {IEEE Conference on Computer Vision and Pattern Recognition (CVPR)}}, 2023.

\bibitem{cheng2024yoloworld}
Tianheng Cheng, Lin Song, Yixiao Ge, Wenyu Liu, Xinggang Wang, and Ying Shan.
\newblock Yolo-world: Real-time open-vocabulary object detection.
\newblock {\em arXiv preprint arXiv:2401.17270}, 2024.

\bibitem{chiang2023vicuna}
Wei-Lin Chiang, Zhuohan Li, Zi~Lin, Ying Sheng, Zhanghao Wu, Hao Zhang, Lianmin Zheng, Siyuan Zhuang, Yonghao Zhuang, Joseph~E Gonzalez, et~al.
\newblock Vicuna: An open-source chatbot impressing gpt-4 with 90\%* chatgpt quality.
\newblock {\em See https://vicuna. lmsys. org (accessed 14 April 2023)}, 2023.

\bibitem{dave2021fixedap}
Achal Dave, Piotr Doll{\'a}r, Deva Ramanan, Alexander Kirillov, and Ross Girshick.
\newblock Evaluating large-vocabulary object detectors: The devil is in the details.
\newblock {\em arXiv preprint arXiv:2102.01066}, 2021.

\bibitem{dosovitskiy2020vit}
Alexey Dosovitskiy, Lucas Beyer, Alexander Kolesnikov, Dirk Weissenborn, Xiaohua Zhai, Thomas Unterthiner, Mostafa Dehghani, Matthias Minderer, Georg Heigold, Sylvain Gelly, Jakob Uszkoreit, and Neil Houlsby.
\newblock An image is worth 16x16 words: Transformers for image recognition at scale.
\newblock In {\em ICLR}, 2020.

\bibitem{GLM}
Zhengxiao Du, Yujie Qian, Xiao Liu, Ming Ding, Jiezhong Qiu, Zhilin Yang, and Jie Tang.
\newblock {GLM:} general language model pretraining with autoregressive blank infilling.
\newblock In {\em {Annual Meeting of the Association for Computational Linguistics (ACL)}}, 2022.

\bibitem{gao2023llamaadapterv2}
Peng Gao, Jiaming Han, Renrui Zhang, Ziyi Lin, Shijie Geng, Aojun Zhou, Wei Zhang, Pan Lu, Conghui He, Xiangyu Yue, et~al.
\newblock Llama-adapter v2: Parameter-efficient visual instruction model.
\newblock {\em arXiv preprint arXiv:2304.15010}, 2023.

\bibitem{girshick2015fastrcnn}
Ross Girshick.
\newblock Fast r-cnn.
\newblock In {\em {IEEE International Conference on Computer Vision (ICCV)}}, 2015.

\bibitem{gupta2019lvis}
Agrim Gupta, Piotr Dollar, and Ross Girshick.
\newblock Lvis: A dataset for large vocabulary instance segmentation.
\newblock In {\em {IEEE Conference on Computer Vision and Pattern Recognition (CVPR)}}, 2019.

\bibitem{gupta2022owdetr}
Akshita Gupta, Sanath Narayan, KJ~Joseph, Salman Khan, Fahad~Shahbaz Khan, and Mubarak Shah.
\newblock Ow-detr: Open-world detection transformer.
\newblock In {\em {IEEE Conference on Computer Vision and Pattern Recognition (CVPR)}}, 2022.

\bibitem{he2017maskrcnn}
Kaiming He, Georgia Gkioxari, Piotr Doll{\'a}r, and Ross Girshick.
\newblock Mask r-cnn.
\newblock In {\em {IEEE International Conference on Computer Vision (ICCV)}}, 2017.

\bibitem{hong2023cogagent}
Wenyi Hong, Weihan Wang, Qingsong Lv, Jiazheng Xu, Wenmeng Yu, Junhui Ji, Yan Wang, Zihan Wang, Yuxiao Dong, Ming Ding, et~al.
\newblock Cogagent: A visual language model for gui agents.
\newblock {\em arXiv preprint arXiv:2312.08914}, 2023.

\bibitem{huang2023tag2text}
Xinyu Huang, Youcai Zhang, Jinyu Ma, Weiwei Tian, Rui Feng, Yuejie Zhang, Yaqian Li, Yandong Guo, and Lei Zhang.
\newblock Tag2text: Guiding vision-language model via image tagging.
\newblock {\em arXiv preprint arXiv:2303.05657}, 2023.

\bibitem{joseph2021ore}
KJ~Joseph, Salman Khan, Fahad~Shahbaz Khan, and Vineeth~N Balasubramanian.
\newblock Towards open world object detection.
\newblock In {\em {IEEE Conference on Computer Vision and Pattern Recognition (CVPR)}}, 2021.

\bibitem{kirillov2023sam}
Alexander Kirillov, Eric Mintun, Nikhila Ravi, Hanzi Mao, Chloe Rolland, Laura Gustafson, Tete Xiao, Spencer Whitehead, Alexander~C Berg, Wan-Yen Lo, et~al.
\newblock Segment anything.
\newblock In {\em {IEEE International Conference on Computer Vision (ICCV)}}, 2023.

\bibitem{krishna2017vg}
Ranjay Krishna, Yuke Zhu, Oliver Groth, Justin Johnson, Kenji Hata, Joshua Kravitz, Stephanie Chen, Yannis Kalantidis, Li-Jia Li, David~A Shamma, et~al.
\newblock Visual genome: Connecting language and vision using crowdsourced dense image annotations.
\newblock {\em {International Journal on Computer Vision (IJCV)}}, 2017.

\bibitem{li2023maskdino}
Feng Li, Hao Zhang, Huaizhe Xu, Shilong Liu, Lei Zhang, Lionel~M Ni, and Heung-Yeung Shum.
\newblock Mask dino: Towards a unified transformer-based framework for object detection and segmentation.
\newblock In {\em {IEEE Conference on Computer Vision and Pattern Recognition (CVPR)}}, 2023.

\bibitem{li2023blip}
Junnan Li, Dongxu Li, Silvio Savarese, and Steven Hoi.
\newblock Blip-2: Bootstrapping language-image pre-training with frozen image encoders and large language models.
\newblock In {\em {International Conference on Machine Learning (ICML)}}, 2023.

\bibitem{li2022coda}
Kaican Li, Kai Chen, Haoyu Wang, Lanqing Hong, Chaoqiang Ye, Jianhua Han, Yukuai Chen, Wei Zhang, Chunjing Xu, Dit-Yan Yeung, et~al.
\newblock Coda: A real-world road corner case dataset for object detection in autonomous driving.
\newblock In {\em {European Conference on Computer Vision (ECCV)}}, 2022.

\bibitem{glip}
Liunian~Harold Li, Pengchuan Zhang, Haotian Zhang, Jianwei Yang, Chunyuan Li, Yiwu Zhong, Lijuan Wang, Lu~Yuan, Lei Zhang, Jenq-Neng Hwang, et~al.
\newblock Grounded language-image pre-training.
\newblock In {\em {IEEE Conference on Computer Vision and Pattern Recognition (CVPR)}}, 2022.

\bibitem{lin2024generateu}
Chuang Lin, Yi~Jiang, Lizhen Qu, Zehuan Yuan, and Jianfei Cai.
\newblock Generative region-language pretraining for open-ended object detection.
\newblock {\em arXiv preprint arXiv:2403.10191}, 2024.

\bibitem{retinanet}
Tsung-Yi Lin, Priya Goyal, Ross Girshick, Kaiming He, and Piotr Doll{\'a}r.
\newblock Focal loss for dense object detection.
\newblock In {\em {IEEE International Conference on Computer Vision (ICCV)}}, 2017.

\bibitem{lin2023sphinx}
Ziyi Lin, Chris Liu, Renrui Zhang, Peng Gao, Longtian Qiu, Han Xiao, Han Qiu, Chen Lin, Wenqi Shao, Keqin Chen, et~al.
\newblock Sphinx: The joint mixing of weights, tasks, and visual embeddings for multi-modal large language models.
\newblock {\em arXiv preprint arXiv:2311.07575}, 2023.

\bibitem{liu2024llava}
Haotian Liu, Chunyuan Li, Qingyang Wu, and Yong~Jae Lee.
\newblock Visual instruction tuning.
\newblock {\em {Neural Information Processing Systems (NeurIPS)}}, 2023.

\bibitem{liu2023groundingdino}
Shilong Liu, Zhaoyang Zeng, Tianhe Ren, Feng Li, Hao Zhang, Jie Yang, Chunyuan Li, Jianwei Yang, Hang Su, Jun Zhu, et~al.
\newblock Grounding dino: Marrying dino with grounded pre-training for open-set object detection.
\newblock {\em arXiv preprint arXiv:2303.05499}, 2023.

\bibitem{liu2021swin}
Ze~Liu, Yutong Lin, Yue Cao, Han Hu, Yixuan Wei, Zheng Zhang, Stephen Lin, and Baining Guo.
\newblock Swin transformer: Hierarchical vision transformer using shifted windows.
\newblock In {\em ICCV}, 2021.

\bibitem{owlv2}
Matthias Minderer, Alexey Gritsenko, and Neil Houlsby.
\newblock Scaling open-vocabulary object detection.
\newblock {\em {Neural Information Processing Systems (NeurIPS)}}, 2023.

\bibitem{peng2023instructiongrit}
Baolin Peng, Chunyuan Li, Pengcheng He, Michel Galley, and Jianfeng Gao.
\newblock Instruction tuning with gpt-4.
\newblock {\em arXiv preprint arXiv:2304.03277}, 2023.

\bibitem{DeFRCN}
Peter Pinggera, Sebastian Ramos, Stefan Gehrig, Uwe Franke, Carsten Rother, and Rudolf Mester.
\newblock Lost and found: detecting small road hazards for self-driving vehicles.
\newblock In {\em International Conference on Intelligent Robots and Systems (IROS)}, 2016.

\bibitem{clip}
Alec Radford, Jong~Wook Kim, Chris Hallacy, Aditya Ramesh, Gabriel Goh, Sandhini Agarwal, Girish Sastry, Amanda Askell, Pamela Mishkin, Jack Clark, et~al.
\newblock Learning transferable visual models from natural language supervision.
\newblock In {\em {International Conference on Machine Learning (ICML)}}, 2021.

\bibitem{ren2015fasterrcnn}
Shaoqing Ren, Kaiming He, Ross Girshick, and Jian Sun.
\newblock Faster r-cnn: Towards real-time object detection with region proposal networks.
\newblock {\em {Neural Information Processing Systems (NeurIPS)}}, 2015.

\bibitem{scheirer2012openset}
Walter~J Scheirer, Anderson de~Rezende~Rocha, Archana Sapkota, and Terrance~E Boult.
\newblock Toward open set recognition.
\newblock {\em {IEEE Transactions on Pattern Recognition and Machine Intelligence (PAMI)}}, 2012.

\bibitem{sun2021sparsercnn}
Peize Sun, Rufeng Zhang, Yi~Jiang, Tao Kong, Chenfeng Xu, Wei Zhan, Masayoshi Tomizuka, Lei Li, Zehuan Yuan, Changhu Wang, et~al.
\newblock Sparse r-cnn: End-to-end object detection with learnable proposals.
\newblock In {\em Proceedings of the IEEE/CVF conference on computer vision and pattern recognition}, pages 14454--14463, 2021.

\bibitem{sun2023eva}
Quan Sun, Yuxin Fang, Ledell Wu, Xinlong Wang, and Yue Cao.
\newblock Eva-clip: Improved training techniques for clip at scale.
\newblock {\em arXiv preprint arXiv:2303.15389}, 2023.

\bibitem{touvron2023llama}
Hugo Touvron, Thibaut Lavril, Gautier Izacard, Xavier Martinet, Marie-Anne Lachaux, Timoth{\'e}e Lacroix, Baptiste Rozi{\`e}re, Naman Goyal, Eric Hambro, Faisal Azhar, et~al.
\newblock Llama: Open and efficient foundation language models.
\newblock {\em arXiv preprint arXiv:2302.13971}, 2023.

\bibitem{wang2023cogvlm}
Weihan Wang, Qingsong Lv, Wenmeng Yu, Wenyi Hong, Ji~Qi, Yan Wang, Junhui Ji, Zhuoyi Yang, Lei Zhao, Xixuan Song, et~al.
\newblock Cogvlm: Visual expert for pretrained language models.
\newblock {\em arXiv preprint arXiv:2311.03079}, 2023.

\bibitem{FSDet}
Xin Wang, Thomas~E. Huang, Joseph Gonzalez, Trevor Darrell, and Fisher Yu.
\newblock Frustratingly simple few-shot object detection.
\newblock In {\em {International Conference on Machine Learning (ICML)}}, 2020.

\bibitem{wang2023detecting}
Zhenyu Wang, Yali Li, Xi~Chen, Ser-Nam Lim, Antonio Torralba, Hengshuang Zhao, and Shengjin Wang.
\newblock Detecting everything in the open world: Towards universal object detection.
\newblock In {\em {IEEE Conference on Computer Vision and Pattern Recognition (CVPR)}}, 2023.

\bibitem{wen2023gpt4v}
Licheng Wen, Xuemeng Yang, Daocheng Fu, Xiaofeng Wang, Pinlong Cai, Xin Li, Tao Ma, Yingxuan Li, Linran Xu, Dengke Shang, et~al.
\newblock On the road with gpt-4v (ision): Early explorations of visual-language model on autonomous driving.
\newblock {\em arXiv preprint arXiv:2311.05332}, 2023.

\bibitem{wu2023exploring}
Jiannan Wu, Yi~Jiang, Bin Yan, Huchuan Lu, Zehuan Yuan, and Ping Luo.
\newblock Exploring transformers for open-world instance segmentation.
\newblock In {\em {IEEE International Conference on Computer Vision (ICCV)}}, 2023.

\bibitem{xiong2023efficientsam}
Yunyang Xiong, Bala Varadarajan, Lemeng Wu, Xiaoyu Xiang, Fanyi Xiao, Chenchen Zhu, Xiaoliang Dai, Dilin Wang, Fei Sun, Forrest Iandola, et~al.
\newblock Efficientsam: Leveraged masked image pretraining for efficient segment anything.
\newblock {\em arXiv preprint arXiv:2312.00863}, 2023.

\bibitem{yao2022detclip}
Lewei Yao, Jianhua Han, Youpeng Wen, Xiaodan Liang, Dan Xu, Wei Zhang, Zhenguo Li, Chunjing Xu, and Hang Xu.
\newblock Detclip: Dictionary-enriched visual-concept paralleled pre-training for open-world detection.
\newblock {\em {Neural Information Processing Systems (NeurIPS)}}, 2022.

\bibitem{yao2024detclipv3}
Lewei Yao, Renjie Pi, Jianhua Han, Xiaodan Liang, Hang Xu, Wei Zhang, Zhenguo Li, and Dan Xu.
\newblock Detclipv3: Towards versatile generative open-vocabulary object detection.
\newblock {\em arXiv preprint arXiv:2404.09216}, 2024.

\bibitem{yuan2024Open-Vocabulary-SAM}
Haobo Yuan, Xiangtai Li, Chong Zhou, Yining Li, Kai Chen, and Chen~Change Loy.
\newblock Open-vocabulary sam: Segment and recognize twenty-thousand classes interactively.
\newblock {\em arXiv preprint arXiv:2401.02955}, 2024.

\bibitem{mobile_sam}
Chaoning Zhang, Dongshen Han, Yu~Qiao, Jung~Uk Kim, Sung-Ho Bae, Seungkyu Lee, and Choong~Seon Hong.
\newblock Faster segment anything: Towards lightweight sam for mobile applications.
\newblock {\em arXiv preprint arXiv:2306.14289}, 2023.

\bibitem{zhang2023llavaground}
Hao Zhang, Hongyang Li, Feng Li, Tianhe Ren, Xueyan Zou, Shilong Liu, Shijia Huang, Jianfeng Gao, Lei Zhang, Chunyuan Li, et~al.
\newblock Llava-grounding: Grounded visual chat with large multimodal models.
\newblock {\em arXiv preprint arXiv:2312.02949}, 2023.

\bibitem{zhang2022glipv2}
Haotian Zhang, Pengchuan Zhang, Xiaowei Hu, Yen-Chun Chen, Liunian Li, Xiyang Dai, Lijuan Wang, Lu~Yuan, Jenq-Neng Hwang, and Jianfeng Gao.
\newblock Glipv2: Unifying localization and vision-language understanding.
\newblock {\em {Neural Information Processing Systems (NeurIPS)}}, 2022.

\bibitem{zhang2023llamaadapter}
Renrui Zhang, Jiaming Han, Chris Liu, Peng Gao, Aojun Zhou, Xiangfei Hu, Shilin Yan, Pan Lu, Hongsheng Li, and Yu~Qiao.
\newblock Llama-adapter: Efficient fine-tuning of language models with zero-init attention.
\newblock {\em arXiv preprint arXiv:2303.16199}, 2023.

\bibitem{zhang2023persam}
Renrui Zhang, Zhengkai Jiang, Ziyu Guo, Shilin Yan, Junting Pan, Xianzheng Ma, Hao Dong, Peng Gao, and Hongsheng Li.
\newblock Personalize segment anything model with one shot.
\newblock {\em arXiv preprint arXiv:2305.03048}, 2023.

\bibitem{zhu2023minigpt}
Deyao Zhu, Jun Chen, Xiaoqian Shen, Xiang Li, and Mohamed Elhoseiny.
\newblock Minigpt-4: Enhancing vision-language understanding with advanced large language models.
\newblock {\em arXiv preprint arXiv:2304.10592}, 2023.

\bibitem{zhu2020deformable}
Xizhou Zhu, Weijie Su, Lewei Lu, Bin Li, Xiaogang Wang, and Jifeng Dai.
\newblock Deformable {DETR:} deformable transformers for end-to-end object detection.
\newblock In {\em {International Conference on Learning Representations (ICLR)}}, 2021.

\bibitem{zhu2024llavaphi}
Yichen Zhu, Minjie Zhu, Ning Liu, Zhicai Ou, Xiaofeng Mou, and Jian Tang.
\newblock Llava-phi: Efficient multi-modal assistant with small language model.
\newblock {\em arXiv preprint arXiv:2401.02330}, 2024.

\end{thebibliography}

\end{document}